\documentclass[10pt,twocolumn,letterpaper]{article}

\usepackage[algorithms]{wacv}      %

\usepackage{graphicx}
\usepackage{amsmath}
\usepackage{amssymb}
\usepackage{booktabs}
\usepackage{multirow}
\usepackage{xcolor}
\usepackage{array}
\usepackage{makecell}
\usepackage{pifont}
\usepackage{enumitem}

\newcommand{\cmark}{\textcolor{green}{\ding{51}}} 
\newcommand{\xmark}{\textcolor{magenta}{\ding{55}}} 

\usepackage[pagebackref,breaklinks,colorlinks]{hyperref}

\usepackage[capitalize]{cleveref}
\crefname{section}{Sec.}{Secs.}
\Crefname{section}{Section}{Sections}
\Crefname{table}{Table}{Tables}
\crefname{table}{Tab.}{Tabs.}

\def\wacvPaperID{88} 
\def\confName{WACV}
\def\confYear{2025}

\begin{document}

\title{Gaussian Déjà-vu: Creating Controllable 3D Gaussian Head-Avatars with Enhanced Generalization and Personalization Abilities}

\author{
Peizhi Yan$^{1}$,
Rabab Ward$^{1}$,
Qiang Tang$^{2}$,
Shan Du$^{3}$\thanks{Corresponding author: Shan Du.\\Supported by funding: GR017752 and GR025942.}\\
$^{1}$University of British Columbia {\tt\small \{yanpz, rababw\}@ece.ubc.ca}\\
$^{2}$Huawei Canada {\tt\small qiang.tang@huawei.com}\\
$^{3}$University of British Columbia (Okanagan) {\tt\small shan.du@ubc.ca}
}
\maketitle

\begin{abstract}
Recent advancements in 3D Gaussian Splatting (3DGS) have unlocked significant potential for modeling 3D head avatars, providing greater flexibility than mesh-based methods and more efficient rendering compared to NeRF-based approaches. Despite these advancements, the creation of controllable 3DGS-based head avatars remains time-intensive, often requiring tens of minutes to hours. To expedite this process, we here introduce the ``Gaussian Déjà-vu" framework, which first obtains a generalized model of the head avatar and then personalizes the result. The generalized model is trained on large 2D (synthetic and real) image datasets. This model provides a well-initialized 3D Gaussian head that is further refined using a monocular video to achieve the personalized head avatar. For personalizing, we propose learnable expression-aware rectification blendmaps to correct the initial 3D Gaussians, ensuring rapid convergence without the reliance on neural networks. Experiments demonstrate that the proposed method meets its objectives. It outperforms state-of-the-art 3D Gaussian head avatars in terms of photorealistic quality as well as reduces training time consumption to at least a quarter of the existing methods, producing the avatar in minutes. Project homepage: \url{https://peizhiyan.github.io/docs/dejavu}
\end{abstract}

\begin{figure}[ht]
    \centering
    \includegraphics[width=1.0\linewidth]{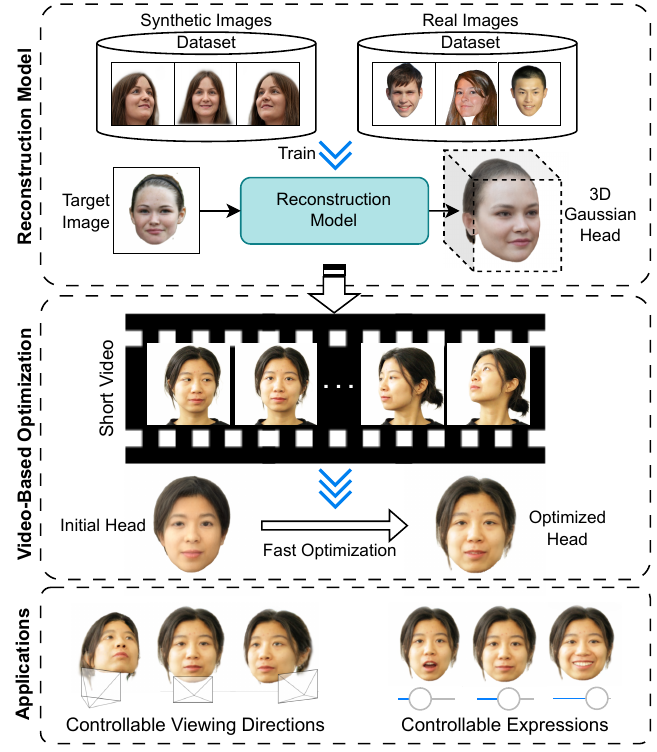}
    \caption{Gaussian Déjà-vu first trains a reconstruction model on large face image datasets and serves as a generalized base. This model initializes the 3D Gaussian head, which is then optimized to personalize the avatar to match the person in the video.}
    \label{fig:teaser}
\end{figure}

\section{Introduction}
\label{sec:intro}

The creation of photorealistic and controllable 3D head avatars has become a popular topic due to their employment in video gaming, VR and AR, filmmaking, telepresence, and many other fields. Three important factors must be considered when creating a 3D head avatar: \textbf{efficiency}, \textbf{quality}, and \textbf{controllability}. Efficiency is related to both the training and the rendering processes. Quality refers to achieving photorealistic results that resemble real human appearances. Controllability involves the ability to easily manipulate facial expressions, head poses, and camera views. Meeting these requirements ensures that the avatars can be effortlessly integrated into various applications. While existing methods may address some of these factors, they often fail to address all. To overcome this, we propose Gaussian Déjà-vu (Déjà-vu), which enables efficient, high-quality, and controllable 3D Gaussian head avatar creation. Table~\ref{table:methods_features} summarizes the features supported by different 3D head avatar creation methods.

\begin{table}[h]
\centering
\scriptsize
\begin{tabular}{rcccc}
\hline
   & \textbf{\makecell{Mesh-\\based}} & \textbf{\makecell{NeRF-\\based}} & \textbf{\makecell{Existing \\ 3DGSs}} & \textbf{Ours}   \\ \hline
\makecell[r]{Single-Image Reconstruction} & \cmark & \cmark & \xmark & \cmark   \\ 
\makecell[r]{Easy Animation}              & \cmark & \xmark & \cmark & \cmark  \\ 
\makecell[r]{Flexible Shape}              & \xmark & \cmark & \cmark & \cmark   \\ 
\makecell[r]{Fast Training on Videos}     & N/A & \xmark & \xmark & \cmark  \\ 
\makecell[r]{Efficient Rendering}         & \cmark & \xmark & \cmark & \cmark \\ \hline
\end{tabular}
\caption{Comparison of supported features.}
\label{table:methods_features}
\end{table}

Mesh-based 3D Morphable Models (3DMMs) have been the foundation of existing 3D head avatars due to their efficiency in rendering and simplicity in animation \cite{blanz2023morphable, paysan20093d, booth20163d, egger20203d}. 3DMMs also support single-image-based reconstruction via either analysis-by-synthesis (fitting-based) \cite{blanz2023morphable, romdhani2003efficient} or learning-based \cite{deng2019accurate, feng2021learning, lin2020towards, kuang2024aware, yan2022neo} methods to allow easy creation of a personal 3D head avatar. Despite their wide acceptance in industries, 3DMMs have a major drawback: the rigidity of mesh topology, making it challenging to model complex parts like hair.

Neural implicit-based methods \cite{mescheder2019occupancy, park2019deepsdf, mildenhall2020nerf}, particularly the Neural Radiance Fields (NeRF) \cite{mildenhall2020nerf} model the 3D scene as a continuous volumetric representation field learned by a neural network, allowing for rendering complex scenes without the need for explicit 3D shapes. NeRF-based 3D face/head methods can be further categorized into conventional NeRF-based methods \cite{gafni2021dynamic, zhuang2022mofanerf, galanakis20233dmm, hong2022headnerf, teotiahq3davatar, chatziagapi2024mi} and StyleGAN+NeRF-based methods \cite{chan2022efficient, zhou2023cips, ma2023otavatar, xu2023omniavatar, sun2023next3d, an2023panohead, kabadayi2024gan, chu2024gpavatar, huang2024g, bao2024geneavatar}. Conventional NeRF-based methods mainly rely on multi-view images/videos as training data, while StyleGAN+NeRF-based methods leverage StyleGAN's network design and learning strategy to enable the training on in-the-wild images. Although NeRF-based methods can provide photorealistic rendering results, the rendering efficiency of NeRF-based methods is still far from practical use, even with some techniques to improve the rendering efficiency \cite{hong2022headnerf, chan2022efficient, xu2023omniavatar, sun2023next3d, teotiahq3davatar, bai2024efficient}. In addition, StyleGAN+NeRF-based methods often encounter flickering issues when used for generating video sequences.

3D Gaussian Splatting (3DGS) is the recent advancement in 3D representation and rendering \cite{kerbl20233d}. It offers a promising way to model 3D head avatars, providing more flexibility than mesh-based methods and also more efficient rendering than NeRF-based methods. The core idea of 3DGS is to use 3D geometry primitives, namely 3D Gaussians, to represent 3D scenes, allowing for smooth and efficient rendering of complex scenes by optimizing the density and attributes of the 3D Gaussians. One important aspect of learning a 3DGS-based head avatar is the initialization of 3D Gaussians. Some works randomly initialize the locations of 3D Gaussians \cite{wang2023gaussianhead, chen2023monogaussianavatar}, while most initialize the locations based on the 3D head mesh to make early-stage learning more efficient \cite{qian2024gaussianavatars, dhamo2023headgas, xiang2023flashavatar, saito2024relightable, lan2023gaussian3diff, xu2024gaussian, luo2024splatface, shao2024splattingavatar, ma20243d, giebenhain2024npga, li2024ggavatar}. Most of these methods focus on the controllability in the animation via training a personalized 3D Gaussian head avatar using the video of a person \cite{wang2023gaussianhead, chen2023monogaussianavatar, qian2024gaussianavatars, dhamo2023headgas, xiang2023flashavatar, saito2024relightable, xu2024gaussian, luo2024splatface, shao2024splattingavatar, ma20243d, giebenhain2024npga, li2024ggavatar}. However, a 3D Gaussian contains more information beyond its location, and only initializing the locations of the 3D Gaussians still leads to slow convergence. Most methods take tens of minutes or even hours to train an animatable 3D Gaussian head avatar for a single person, significantly limiting their wide application. We believe that a single-image-based reconstruction model can provide a good initialization of the 3D Gaussians. Nevertheless, existing 3DGS-based animatable head avatar creation methods do not support using a single 2D image.

In this work, we propose the ``Gaussian Déjà-vu" framework (Déjà-vu) to efficiently create a controllable 3DGS-based head avatar. We parameterize the 3D Gaussians into a specialized UV map representation (a 2D representation of a 3D object's surface), which we call the UV Gaussian map, aligned with the mesh-based FLAME model \cite{li2017learning} to enable shape initialization and controllable animation. We train a reconstruction network to generate UV offsets that will be used to correct the FLAME-initialized 3D Gaussians \cite{xiang2023flashavatar} related to the single-image-based reconstruction task. The training primarily uses a synthetic dataset with a large number of identities, each captured from multiple viewpoints. We propose a training scheme that randomly pairs the input and target views of the same identity (person), along with a corresponding multi-view consistency regularization loss to enhance the 3D consistency. To further improve the generalization ability, we also fine-tune the network on real images. We only need to train the network once.

We first use our trained single-image-based reconstruction model to provide an initial 3D Gaussian head and then propose a rectification approach to enable further optimization of the head using a monocular video of the intended person. This rectification approach eliminates the need for a neural network by introducing learnable UV rectification maps, which are additional offsets to the initial 3D Gaussians. We propose expression-aware UV rectification blendmaps (blendmaps) to be learned using the given video. Different from 3D Gaussian blendshapes methods \cite{dhamo2023headgas, ma20243d}, our proposed blendmaps only serve to rectify the initial 3D Gaussians, thus the optimization is faster than training a set of blendshapes from scratch. Due to our efficient design, Déjà-vu achieves a real-time animation speed of 220 frames-per-second (fps) at a resolution of $512 \times 512$ and a faster training speed than existing methods, reducing the training time to at least a quarter. Figure~\ref{fig:teaser} provides a brief overview of the proposed framework.

In summary, our contributions are as follows:
\begin{itemize}[nosep, left=0pt]
    \item We propose Déjà-vu, a framework that first develops a single-image-based 3D Gaussian head reconstruction model to initialize a 3D Gaussian head, and further optimize the head using monocular videos to appear real and gain personalization. This approach significantly reduces the training time on videos and produces state-of-the-art quality;
    \item We propose expression-aware UV rectification blendmaps, which allow easy control of facial animation without the need for a neural network;
    \item Our training strategy for the single-image-based reconstruction model involves the use of both synthetic and real images as training data, and also view-consistency regularization, which enhances generalization. The reconstruction model provides a robust initialization for subsequent video-based optimization. To our knowledge, this is the first single-image-based reconstruction model for the 3D Gaussian head.
\end{itemize}

\section{Related Works}
\label{sec:relate}

\subsection{3D Morphable Models in 3D Face Modeling}

Over the past half-century, mesh-based methods have been the mainstream in 3D face modeling and animation \cite{deng2008computer}. A straightforward way to synthesize varying 3D faces is by linearly blending multiple template 3D faces, a method known as blendshapes \cite{vetter1998estimating}. Building on the concept of blendshapes, 3D Morphable Models (3DMMs) compress the 3D face shapes and textures from scanned data to principal components and allow the synthesis of a new 3D face from those principal components with given blending coefficients \cite{blanz2023morphable}. Earlier 3DMMs have only covered the facial region (including neck and ears) \cite{blanz2023morphable, paysan20093d, booth20163d}, while more advanced versions such as FLAME \cite{li2017learning} model have addressed the entire head except for hair. Many 3DMM-based single-image 3D face reconstruction methods estimate the 3DMM coefficients to reconstruct the 3D face \cite{deng2019accurate, feng2021learning, lin2020towards, kuang2024aware}. To leverage the spatial capabilities of convolutional neural networks, some reconstruction methods utilize UV maps as a middle representation \cite{feng2018joint, zhu2020reda, feng2021learning}. The UV maps are 2D representations of the 3D model with predefined 3D-2D correspondences that facilitate the application of textures and detailed surface information onto the 3D geometry \cite{foley1996computer}.

\subsection{NeRF-Based 3D Face and Head Models}


The original Neural Radiance Fields method (NeRF) employs a neural network to represent the static 3D scene learned from multi-view images \cite{mildenhall2020nerf}. MoFaNeRF adapts NeRF to model the human face with parametric control by incorporating facial codes akin to 3DMM coefficients as additional network inputs \cite{zhuang2022mofanerf}. HeadNeRF enhances rendering speed through a coarse-to-fine strategy, utilizing NeRF to render low-resolution feature maps and a 2D neural rendering network for producing high-resolution head images \cite{hong2022headnerf}. 3DMM-RF leverages 3DMM-based synthetic face images for training, achieving more accurate fitting over the facial region \cite{galanakis20233dmm}. Inspired by the success of 2D StyleGANs \cite{karras2020analyzing, karras2021alias} in generating photorealistic face images, StyleNeRF integrates a style-based network architecture into NeRF and trains the network through adversarial learning as in StyleGAN \cite{gu2022stylenerf}. EG3D further refines the network architecture by forming a tri-plane representation for volume rendering of the raw image \cite{chan2022efficient}. Due to efficiency considerations, the tri-plane generally has low resolution, prompting the use of a 2D super-resolution module to enhance the final image resolution. This tri-plane NeRF scheme is now widely adopted by numerous 3D-aware head models \cite{ma2023otavatar, xu2023omniavatar, sun2023next3d, an2023panohead, kabadayi2024gan, chu2024gpavatar}. Although StyleGAN+NeRF methods generate photorealistic rendered head images, they can introduce artifacts, such as noticeable flickering during animations. Moreover, these models do not support the customization of the reconstructed head based on video data of a person, limiting their applicability in more personalized head avatar scenarios. In contrast, the Déjà-vu framework we propose here produces an explicit 3D Gaussian-based head avatar that can be rendered directly through 3DGS \cite{kerbl20233d} and optimized using video data to enhance personalization.


\subsection{3D Gaussian-Based Head Avatars}

Like the original NeRF, the 3D Gaussian Splatting (3DGS) is designed to represent and render static 3D scenes \cite{kerbl20233d}. To achieve controllability, 3DGS-based head avatar methods typically employ 3DMM. GaussianAvatars introduces a technique to rig the 3D Gaussians to the FLAME model, where the Gaussians are anchored to the triangle facets of the mesh \cite{qian2024gaussianavatars}. Similarly, SplattingAvatar proposes a method to embed 3D Gaussians into any head mesh model \cite{shao2024splattingavatar}. SplatFace aligns the 3D Gaussians with the underlying head mesh through a non-rigid process and jointly optimizes both the Gaussians and the mesh \cite{luo2024splatface}. Numerous 3DGS head avatar methods leverage neural rendering to achieve photorealism by incorporating high-dimensional latent features into each 3D Gaussian and utilizing neural networks to decode these features to produce the final images \cite{dhamo2023headgas, xu2024gaussian, giebenhain2024npga, li2024ggavatar, wang2023gaussianhead, lan2023gaussian3diff, barthel2024gaussian, saito2024relightable}. For instance, GaussianHead and Gaussian Splatting Decoder employ a tri-plane representation to learn the 3D Gaussian latent features \cite{wang2023gaussianhead, barthel2024gaussian}. HeadGaS utilizes a Multi-layer Perceptron (MLP) network to decode the latent features of each 3D Gaussian into color and alpha blending weights \cite{dhamo2023headgas}. Additionally, NPGA and Gaussian Head Avatar initially render a 2D feature map using a modified 3DGS, which is then decoded into an image through 2D neural rendering \cite{giebenhain2024npga, xu2024gaussian}. However, these neural-rendering-based methods deviate from the simplicity of traditional 3DGS, resulting in reduced compatibility with existing 3D engines. In addition, some methods employ a ``canonical + deformation" strategy which leverages a neural network, typically an MLP, to estimate the deformation for each 3D Gaussian \cite{xiang2023flashavatar, chen2023monogaussianavatar, li2024ggavatar}. Approaches like HeadGaS and 3D Gaussian Blendshapes integrate the concept of blendshapes by learning a set of 3D Gaussians as expression bases \cite{dhamo2023headgas, ma20243d}. 

Despite the success of existing 3DGS-based head avatar methods, this work takes us one step further, introducing the first single-image-based reconstruction of a 3D Gaussian head avatar, which provides robust initialization for our video-based optimization. Our 3D Gaussian head avatar aligns with the original 3DGS and does not require a neural network for animation, ensuring compatibility and ease of integration with standard 3D engines.

\section{Method}
\label{sec:method}


\subsection{Background: 3D Gaussian Splatting}

The 3D Gaussian Splatting (3DGS) presents a 3D Gaussians-based 3D representation and its corresponding splatting-based rendering process, allowing for differentiable and efficient rendering \cite{kerbl20233d}. 3DGS modifies the Probability Density Function (PDF) of a multivariate Gaussian distribution to define the impact of a 3D Gaussian $\mathbf{g}_i$ on a given 3D location $\mathbf{x} \in \mathbb{R}^3$:
\begin{equation}
    G_i(\mathbf{x}) = e^{-\frac{1}{2} (\mathbf{x} - \mathbf{\mu}_{i})^{\top} \mathbf{\Sigma}_{i}^{-1} (\mathbf{x} - \mathbf{\mu}_{i})},
    \label{eq:3dgaussian}
\end{equation}
where $\mathbf{\mu}_{i} \in \mathbb{R}^{3}$ is the mean (center of the 3D Gaussian) and $\mathbf{\Sigma}_i \in \mathbb{R}^{3 \times 3}$ is the covariance matrix. To ensure the covariance matrix $\mathbf{\Sigma}_i$ to be positive semi-definite to have physical meaning, 3DGS computes the $\mathbf{\Sigma}_i$ based on a scaling matrix $S_i \in \mathbb{R}^{3 \times 3}$ and a rotation matrix $R_i \in \mathbb{R}^{3 \times 3}$:
\begin{equation}
    \mathbf{\Sigma}_i = R_{i} S_{i} S_{i}^{\top} R_{i}^{\top}.
    \label{eq:covariance}
\end{equation}
This representation is analogous to defining the scaling and rotation of an ellipsoid \cite{kerbl20233d}. Further, the scaling and rotation matrices can be derived by a scaling vector $\mathbf{s}_i \in \mathbb{R}^{3}$ and a quaternion $\mathbf{q}_i = (1, \mathbf{r}_i)$, where $\mathbf{r}_i \in \mathbb{R}^{3}$ defines the three rotation angles. The 3D Gaussian $\mathbf{g}_{i}$ is then parameterized by $\mathbf{g}_{i} = \{ \mathbf{\mu}_{i}, \mathbf{s}_i, \mathbf{r}_i, \alpha_i, \mathbf{a}_i \}$, where $\alpha_i$ is the alpha blending weight, and $\mathbf{a}_i$ is the vector of Spherical Harmonic (SH) coefficients for computing the view-dependent RGB color \cite{fridovich2022plenoxels}. The 3D Gaussians are projected to 2D for rendering. The image formation model of 3DGS mirrors the volume rendering techniques utilized in NeRF \cite{drebin1988volume, mildenhall2020nerf}:
\begin{equation}
    \mathbf{C}(p) = \sum_{i \in \mathcal{N}_{p}} \mathbf{c}_i^{p} \alpha_{i}  \prod_{j=1}^{i-1}(1 - \alpha_{j}),
    \label{eq:rendering}
\end{equation}
where $p$ is the 2D-pixel location on the plane onto which 3D Gaussians are projected. Here, $\mathbf{C}(p)$ represents the final color of the pixel, $\mathcal{N}_{p}$ is an ordered set of the 2D projections of the 3D Gaussians that cover the pixel, and $\mathbf{c}_i^{p}$ is the computed color of the $i^{th}$ 2D projection at location $p$ \cite{yifan2019differentiable}.


\begin{figure*}[ht]
    \centering
    \includegraphics[width=1.0\linewidth]{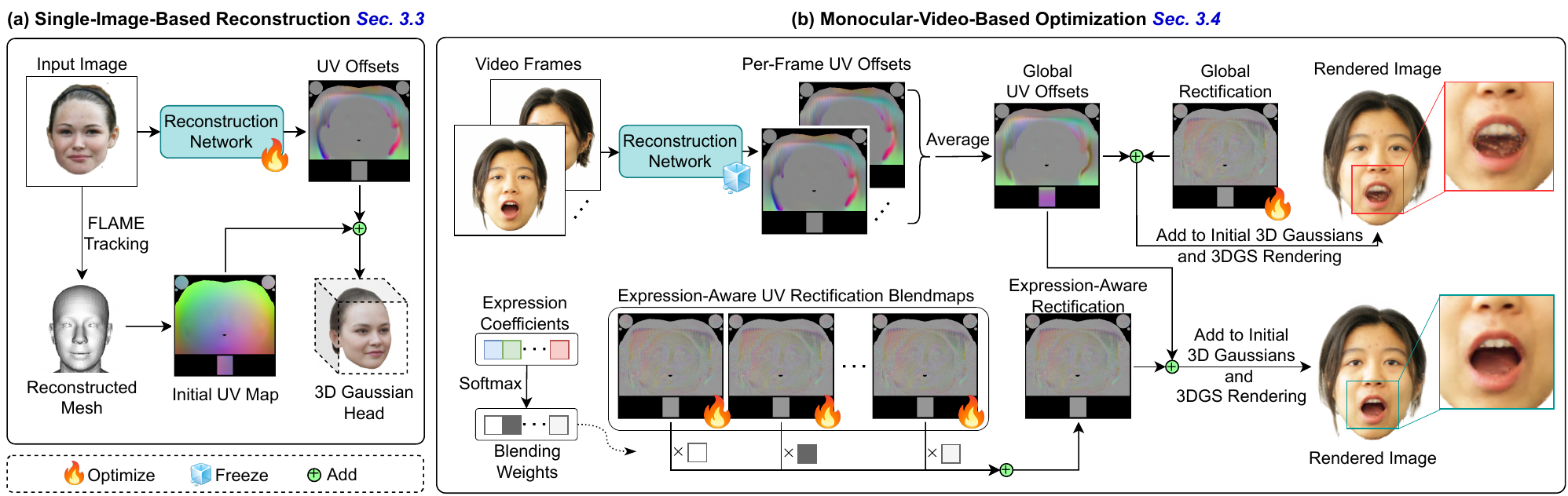}
    \caption{Detailed flowcharts for our single-image-based reconstruction (a) and monocular-video-based further optimization (b) processes. }
    \label{fig:framework}
\end{figure*}

\subsection{Our UV Gaussian Map Representation}
\label{subsec:uv_gaussian_map}

Our 3D Gaussian head avatar is based on FLAME \cite{li2017learning} to gain controllability. We denote $F$ as the FLAME shape model. The 3D head mesh can be constructed through: \( m = F(\beta_{id}, \beta_{exp}, \beta_{jaw}) \),
where $\beta_{id} \in \mathbb{R}^{100}$, $\beta_{exp} \in \mathbb{R}^{50}$, and $\beta_{jaw} \in \mathbb{R}^{3}$ are the coefficients of identity, expression, and jaw pose respectively. We modify the original UV correspondences of FLAME to remove the neck and add the mouth interior shape (see supplementary material). 

We define a set of UV maps with height and width of $H$ and $W$ to represent the parameters of 3D Gaussians. These UV maps include $U_{\mu}$ for the 3D Gaussian positions, $U_{\mathbf{s}}$ for the scales, $U_{\mathbf{r}}$ for the rotations, $U_{\alpha}$ for the alpha blending weights, and $U_{\mathbf{a}}$ for the SH coefficients. Given the relatively simple surface materials of the human head, we do not consider complex anisotropic reflections and refractions. Therefore, we use SHs with a degree of zero to represent RGB colors uniformly across all directions. The number of channels in $U_{\mathbf{a}}$ is thus equals to three. Our UV Gaussian map $U \in \mathbb{R}^{H \times W \times 13}$ is created by channel-wise concatenation of the above set of UV maps. We mask out the invalid pixels (undefined locations in the UV map) in the UV Gaussian map, and each valid pixel corresponds to a 3D Gaussian. Therefore the total number of 3D Gaussians is less than $H * W$. 

To prepare an initial UV Gaussian map $U_{\text{init}}$, we rasterize the mesh shape $m$ to initialize the Gaussian positions $U_{\mu}$. We empirically initialize the values of alpha weights in $U_{\alpha}$ to 0.1 and the values of scales in $U_{\mathbf{s}}$ to $-8.3533$ (note that 3DGS applies exponential activation on the scales, to ensure the final scales are positive). All the other UV maps are initialized with zeros.

\subsection{Single-Image-Based Reconstruction}

Our single-image-based 3D Gaussian head reconstruction utilizes a convolutional encoder-decoder network to estimate the offsets to the initialized 3D Gaussians, see Figure~\ref{fig:framework}~(a). The initial 3D Gaussian head is defined by using the initial UV Gaussian map, as detailed in Section~\ref{subsec:uv_gaussian_map}. We implemented a FLAME tracking method to derive the FLAME coefficients and the camera pose given an image (see our homepage). Our goal is to estimate a set of offsets, referred to as UV offsets $\Delta U$, which adjust the initial UV Gaussian map $U_{\text{init}}$. The encoder processes a head image $I$ to produce low-resolution feature maps, which are subsequently input into the decoder. Additionally, the decoder incorporates the initial UV position map $U_{\mu}$ to enhance adaptation to the UV map layout. The corrected UV Gaussian map for the single-image-reconstruction phase is computed as $U_{\text{SIR}} = U_{\text{init}} + \Delta U$. Using the estimated camera pose for $I$, we render $U_{\text{SIR}}$ through 3DGS, with the resulting image denoted by $I'_{\text{SIR}}$. The reconstruction loss, which quantifies the discrepancy between $I$ and $I'_{\text{SIR}}$, is computed as follow:
\begin{equation}
    \mathcal{L}_{\text{photo}} = \lambda_{\text{RGB}}\mathcal{L}_{\text{RGB}} + \lambda_{\text{LPIPS}}\mathcal{L}_{\text{LPIPS}} +
    \lambda_{\text{SSIM}}\mathcal{L}_{\text{SSIM}},
    \label{eq:photo_loss}
\end{equation}
where $\mathcal{L}_{\text{RGB}}$ is the pixel-level L1 loss,  $\mathcal{L}_{\text{LPIPS}}$ is the perceptual loss \cite{zhang2018unreasonable}, $\mathcal{L}_{\text{SSIM}}$ is the differentiable SSIM-based loss \cite{wang2003multiscale}, and the $\lambda$s are corresponding loss term weights.

To ensure that the positions of the 3D Gaussians do not deviate significantly from the original reconstructed FLAME shape, we introduce a position regularization term, denoted as $\mathcal{L}_{\mu}$:
\begin{equation}
    \mathcal{L}_{\mu} = \| \Delta U_{\mu} \circ M_{\mu} \|_{2}, 
    \label{eq:reg_position}
\end{equation}
where $\Delta U_{\mu}$ is the network estimated UV position offsets map, $\circ$ denotes the element-wise Hadamard product operation, and $M_{\mu}$ is a predefined weights map. This map allocates more weight to the facial region and less to the scalp, ensuring that the facial area more closely conforms to the reconstructed FLAME shape, while permitting the hair, represented by the 3D Gaussians originally on the scalp, to maintain some positional variability. Similarly, we define the scale regularization term, denoted as $\mathcal{L}_{\mathbf{s}}$ to regularize the size of the Gaussians from growing excessively:
\begin{equation}
    \mathcal{L}_{\mathbf{s}} = \| \Delta U_{\mathbf{s}} \|_{2}. 
    \label{eq:reg_scale}
\end{equation}

We introduce a view consistency regularization loss, $\mathcal{L}_{\text{view}}$, to ensure stability across different views of the same identity:
\begin{equation}
    \mathcal{L}_{\text{view}} = \sum_{\Delta U \in \mathcal{B}} \| \Delta U - \Delta \bar{U}_{\mathcal{B}} \|_{2} / |\mathcal{B}|, 
    \label{eq:reg_view}
\end{equation}
where $\mathcal{B}$ represents a set of UV offsets from multi-view input images for the same identity, and $\Delta \bar{U}_{\mathcal{B}}$ denotes the mean UV offsets within the set. To further enhance view consistency, we also randomly pair the input and target images from the same batch of multi-view images belonging to the same identity during training. The overall loss is:
\begin{equation}
    \mathcal{L} = \mathcal{L}_{\text{photo}} + \lambda_{\mu}\mathcal{L}_{\mu} + \lambda_{\mathbf{s}}\mathcal{L}_{\mathbf{s}} + \lambda_{\text{view}}\mathcal{L}_{\text{view}}. 
    \label{eq:total_loss}
\end{equation}

\subsection{Monocular-Video-Based Optimization}

Using a monocular video of a person, we apply our single-image-based reconstruction to each frame and then average the estimated UV offsets to obtain the mean UV offsets for the video, denoted as $\Delta \bar{U}$. Merely using $U_{\text{init}} + \Delta \bar{U}$ to form the reconstructed 3D Gaussian head may result in a lack of personalization. To address this, we propose a two-stage optimization method aimed at further enhancing the personalization and quality of the reconstructed 3D Gaussian head. See Figure~\ref{fig:framework} (b).

In the first stage (Figure~\ref{fig:framework} (b) top), we define the global UV rectification, $\Delta U_{\text{global}}$ as a tensor in $\mathbb{R}^{H \times W \times 13}$, which is initially set to zeros and made learnable. The UV Gaussian map, after applying this global UV rectification, is expressed as:
\begin{equation}
    U_{\text{global}} = U_{\text{init}} + \Delta \bar{U} + \Delta U_{\text{global}}.
    \label{eq:global_rectify}
\end{equation}
The objective of the first stage is to optimize $\Delta U_{\text{global}}$ to minimize $\mathcal{L}$ in Equation~\ref{eq:total_loss} (excluding $\mathcal{L}_{\text{view}}$). After the first stage of optimization, the resulting 3D Gaussian head already closely resembles the person in the video. However, relying solely on $\Delta U_{\text{global}}$ can still lead to artifacts during animation, as this approach does not account for variations in facial expressions. 

To address this limitation, we introduce a second stage (Figure~\ref{fig:framework} (b) bottom) that optimizes the UV rectification blendmaps $\Delta U_{\text{blend}} \in \mathbb{R}^{D \times H \times W \times 13}$, where $D$ denotes the number of blending weights and $\Delta U_{\text{blend}}^{i}$ denotes the $i^{th}$ blendmap. Our blending weights $b \in \mathbb{R}^{D}$ are derived from the estimated FLAME expression coefficients $\beta_{exp}$:
\begin{equation}
    b = softmax( \beta_{exp} ),
    \label{eq:blending_weights}
\end{equation}
where $b_{i}$ indicates the $i^{th}$ blending weight. Utilizing the softmax function ensures the blending weights are positive values and have a sum of exactly 1. We initialize the blendmaps with the $\Delta U_{\text{global}}$ learned in the first stage. The final UV Gaussian map after rectification is expressed as:
\begin{equation}
    U_{\text{blend}} = U_{\text{init}} + \Delta \bar{U} + \sum_{i=1}^{D} b_{i} \Delta U_{\text{blend}}^{i}.
    \label{eq:blend_rectify}
\end{equation}
We optimize $\Delta U_{\text{blend}}$ with the same objective as in the first stage.

\section{Experiments}
\label{sec:experiments}

\subsection{Datasets}

We prepare two datasets for training our single-image-based reconstruction model. The first is a synthetic dataset created using the 3D-aware face image generation model, PanoHead \cite{an2023panohead}, which synthesized images for 18,000 identities across 25 pre-defined camera views (detailed in the supplementary material). The second dataset comprises real-face images from the FFHQ dataset \cite{karras2019style} to improve the robustness of our network on real images. We filtered out images where faces are obscured by hats or clothing, or of low quality, resulting in 38,000 usable images. Of these, the first 1,000 images are set aside for evaluation, while the remaining 37,000 images are used for training. For video-based optimization, we utilize the dataset from PointAvatar \cite{zheng2023pointavatar}, which includes monocular video clips of three subjects \cite{zheng2023pointavatar}. For each subject, we allocate the first 90\% of the frames from each video clip for training purposes, reserving the remaining 10\% for evaluation.

\subsection{Implementation Details}

We set the resolution of our UV Gaussian map to $320 \times 320$ in our experiment. The actual number of Gaussians is 74,083. The loss term weights are set as follows: $\lambda_{\text{RGB}} = 1.0$, $\lambda_{\text{LPIPS}} = 0.5$, $\lambda_{\text{SSIM}} = 0.05$, $\lambda_{\mu} = 0.5$, $\lambda_{\mathbf{s}} = 5\times 10^{-5}$, and $\lambda_{\text{view}} = 0.1$. The rendering resolution is $512 \times 512$.

During the training of the single-image-based reconstruction network, each training step starts with training on a random batch from the synthetic dataset at a higher learning rate. This is followed by fine-tuning on a random batch from the real dataset at a reduced learning rate, specifically one-tenth of that used for the synthetic dataset. It's important to note that the $\mathcal{L}_{\text{view}}$ term is omitted during fine-tuning on the in-the-wild dataset, as multi-view images are not available. The training batch size is set to 16.

For video-based optimization, the distribution of steps between the first and second stages is split into a 3:7 ratio. The batch size for each step is 8. We utilize only the first ten expression coefficients to compute the blending weights, where $D = 10$.

All the experiments are performed on one Nvidia A6000 GPU (48GB VRAM). For additional details, please refer to the supplementary material.

\subsection{Single-Image-Based Reconstruction Results}

\begin{figure}[h]
    \centering
    \includegraphics[width=.8\linewidth]{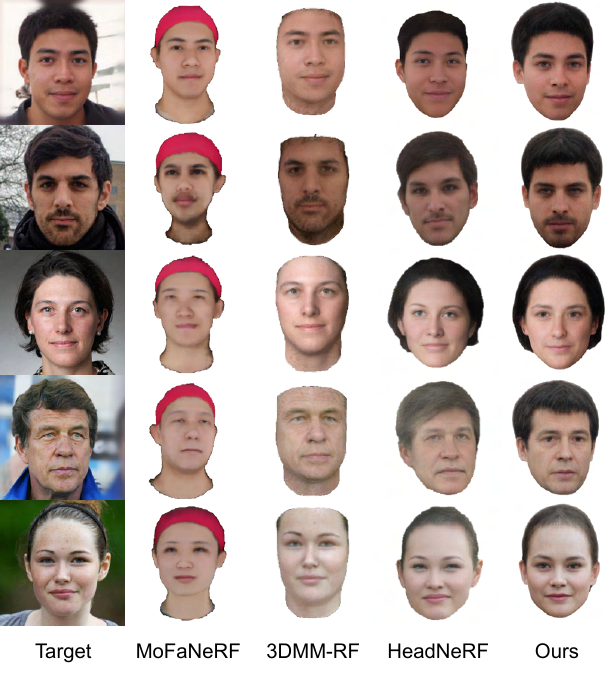}
    \caption{Qualitative comparison results on single-image-based reconstruction.}
    \label{fig:single-image-recon}
\end{figure}

We compare the performance of our single-image-based reconstruction model with leading NeRF-based parametric face/head models, MoFaNeRF \cite{zhuang2022mofanerf}, HeadNeRF \cite{hong2022headnerf}, and 3DMM-RF \cite{galanakis20233dmm}, due to their aligned goal with us on training a general model for face and head reconstruction. We exclude comparisons with StyleGAN-based or diffusion-based methods, as these primarily focus on image synthesis rather than creating a general head model. Since MoFaNeRF and 3DMM-RF only represent the facial region, we compare only the facial region with these methods. For a fair comparison, we employ the same set of 550 testing samples used by 3DMM-RF, extracted from the CelebAMask-HQ dataset \cite{CelebAMask-HQ}. We assess reconstruction performance using several metrics: pixel-level L1 distance, LPIPS \cite{zhang2018unreasonable}, Structural Similarity Index (SSIM) \cite{wang2004image}, and Peak Signal-to-Noise Ratio (PSNR). Detailed results of these comparisons are presented in Table~\ref{tab:compare-celeb}. Additionally, we conduct tests on a separate set of 1,000 images from the FFHQ dataset \cite{karras2019style}, with outcomes detailed in Table~\ref{tab:compare-ffhq}. Our method demonstrates competitive performance, frequently outperforming these benchmarks in most metrics. Figure~\ref{fig:single-image-recon} presents qualitative comparisons, showing that our method achieves a visual quality closely resembling the source image.

\begin{table}[h]
    \footnotesize
    \centering
    \setlength{\tabcolsep}{4.5pt} 
    \begin{tabular}{lccccc}
    \Xhline{2\arrayrulewidth}
        \textbf{Method} & \textbf{Region} & \textbf{L1} $\downarrow$ & \textbf{LPIPS} $\downarrow$ & \textbf{SSIM} $\uparrow$ & \textbf{PSNR} $\uparrow$ \\\hline

        MoFaNeRF \cite{zhuang2022mofanerf} & \multirow{4}{*}{Facial} & 0.273 & 0.442 & 0.910 & 14.713 \\ 
        
        3DMM-RF* \cite{galanakis20233dmm} &  & 0.216 & - & \textbf{0.956} & - \\ 


        HeadNeRF \cite{hong2022headnerf} &  & 0.072 & 0.113 & 0.940 & 24.937 \\ 

        Ours     &  & \textbf{0.033} & \textbf{0.061} & 0.935 & \textbf{28.755} \\ \hline


        HeadNeRF \cite{hong2022headnerf} & \multirow{2}{*}{Head} & 0.354 & 0.431 & 0.693 & 15.429 \\

        Ours     &  & \textbf{0.125} & \textbf{0.211} & \textbf{0.740} & \textbf{20.525} \\
        \Xhline{2\arrayrulewidth}
        \multicolumn{6}{l}{* indicates results reported from the original paper (not open-source).}\\ 
        \multicolumn{6}{l}{- indicates the value is missing or not comparable.}\\
    \end{tabular}
    \caption{Reconstruction performances (CelebAMask-HQ).}
    \label{tab:compare-celeb}
\end{table}

\begin{table}[h]
    \footnotesize
    \centering
    \setlength{\tabcolsep}{4.5pt} 
    \begin{tabular}{lccccc} 
    \Xhline{2\arrayrulewidth}
        \textbf{Method} & \textbf{Region} & \textbf{L1} $\downarrow$ & \textbf{LPIPS} $\downarrow$ & \textbf{SSIM} $\uparrow$ & \textbf{PSNR} $\uparrow$ \\\hline

        MoFaNeRF \cite{zhuang2022mofanerf}  & \multirow{3}{*}{Facial} & 0.251 & 0.416 & 0.720 & 15.084 \\
        

        HeadNeRF \cite{hong2022headnerf}  & & 0.087 & 0.151 & 0.920 & 23.858 \\

        Ours     &  & \textbf{0.040} & \textbf{0.071} & \textbf{0.922} & \textbf{28.030} \\ \hline


        HeadNeRF \cite{hong2022headnerf} & \multirow{2}{*}{Head} & 0.211 & 0.310 & 0.798 & 18.794 \\

        Ours     &  & \textbf{0.088} & \textbf{0.155} & \textbf{0.813} & \textbf{23.184} \\ 
    \Xhline{2\arrayrulewidth}

        
    \end{tabular}
    \caption{Reconstruction performances (FFHQ).}
    \label{tab:compare-ffhq}
\end{table}

\begin{figure}[h]
    \centering
    \includegraphics[width=0.8\linewidth]{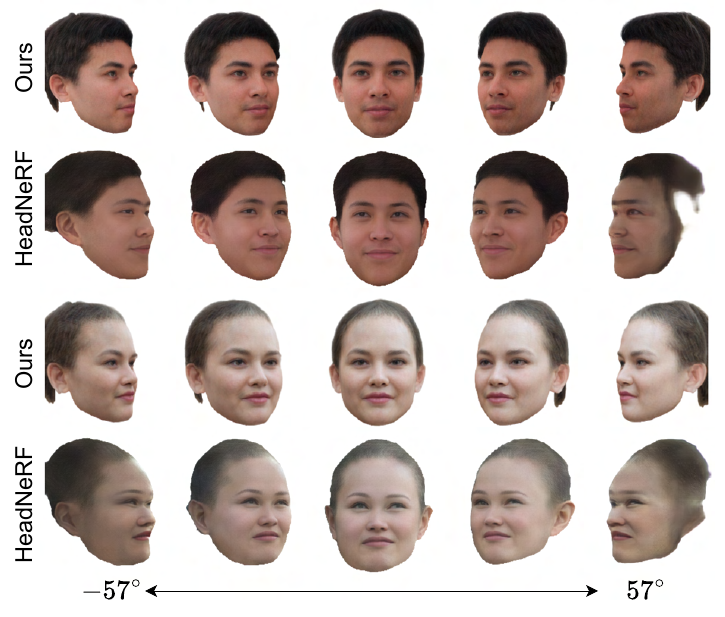}
    \caption{Qualitative comparison with HeadNeRF across varying viewing angles. Our method works even at extreme angles.}
    \vspace{-0.2cm}
    \label{fig:multi-view}
\end{figure}

We compare our method with HeadNeRF to showcase the superior 3D view consistency of our 3D Gaussian-based head representation. Figure~\ref{fig:multi-view} presents comparisons, illustrating the rendering of the reconstructed head from various viewing angles. As shown, our method consistently outperforms HeadNeRF in maintaining view consistency, attributable to the explicit 3D structure represented by the 3D Gaussians. Notably, at extreme viewing angles, where HeadNeRF’s renderings show significant deterioration, our 3D Gaussian-based head continues to maintain high-quality visual outputs.

\subsection{Monocular-Video-Based Optimization Results}

\begin{table*}[h]
\centering
\footnotesize
\renewcommand{\arraystretch}{1.0} 
\setlength{\tabcolsep}{4.9pt} 
\begin{tabular}{l|c|ccc|ccc|ccc}
    \Xhline{2\arrayrulewidth}
    \multicolumn{1}{c|}{} & Training & \multicolumn{3}{c|}{\textit{Subject 1}} & \multicolumn{3}{c|}{\textit{Subject 2}} & \multicolumn{3}{c}{\textit{Subject 3}} \\  
    \multicolumn{1}{c|}{} & Time & LPIPS $\downarrow$ & SSIM $\uparrow$ & PSNR $\uparrow$ & LPIPS $\downarrow$ & SSIM $\uparrow$ & PSNR $\uparrow$ & LPIPS $\downarrow$ & SSIM $\uparrow$ & PSNR $\uparrow$ \\ \hline
    \multirow{2}{*}{GaussianAvatars (CVPR'24) \cite{qian2024gaussianavatars}} & 3 hrs & 0.155 & 0.877 & 22.650 & 0.141 & 0.932 & 28.672 & 0.135 & 0.894 & 22.152 \\
     & 6 hrs & 0.155 & 0.879 & 22.999 & 0.141 & 0.933 & 29.094 & 0.134 & 0.895 & 22.101 \\
    \hline
    \multirow{2}{*}{FlashAvatar (CVPR'24) \cite{xiang2023flashavatar}} & 30 mins & 0.104 & 0.867 & 23.565 & 0.091 & 0.917 & 28.447 & 0.096 & 0.872 & 22.028 \\ 
     & 60 mins & 0.097 & 0.875 & 23.767 & 0.091 & 0.917 & 28.561 & 0.091 & 0.874 & 21.973 \\ 
    \hline
    Ours (global rectification only) & 8 mins & 0.082 & 0.883 & 23.157 & 0.073 & 0.931 & 29.083 & 0.077 & 0.883 & 21.487 \\ 
    Ours & 8 mins & \textbf{0.073} & \textbf{0.899} & \textbf{24.277} & \textbf{0.070} & \textbf{0.936} & \textbf{29.455} & \textbf{0.067} & \textbf{0.899} & \textbf{22.342} \\ 
    \Xhline{2\arrayrulewidth}
\end{tabular}
\caption{Quantitative comparison with state-of-the-art 3D head avatar methods.}
\label{table:imavatar_compare}
\vspace{-0.25cm}
\end{table*}

\begin{figure}[h]
    \centering
    \includegraphics[width=0.85\linewidth]{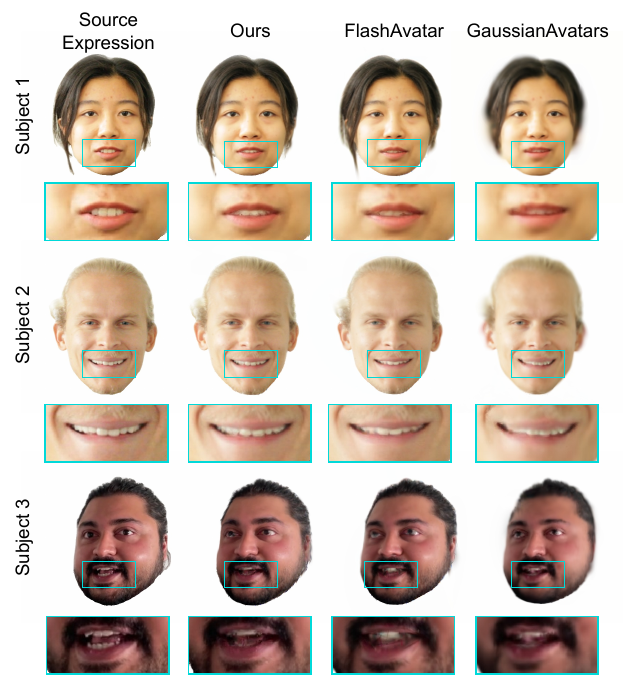}
    \caption{Expression reconstruction results.}
    \label{fig:imavatar_sota}
\end{figure}

In the monocular-video-based optimization, we compare our method with the state-of-the-art 3D Gaussian head avatar methods, GaussianAvatars \cite{qian2024gaussianavatars} and FlashAvatar \cite{xiang2023flashavatar}, both of which are based on the FLAME model. To ensure fairness, we use the same FLAME tracking results to obtain expression coefficients, the underlying mesh, and the 6DOF camera pose. Additionally, as FlashAvatar also utilizes a fixed number of 3D Gaussians, we use the same amount of 74,083 3D Gaussians as in our method. GaussianAvatars incorporates dynamic density control from 3DGS, but each learned head model has around 60,000 to 100,000 3D Gaussians, which is similar to ours. Table~\ref{table:imavatar_compare} shows the quantitative comparison results. Our method outperforms both GaussianAvatars and FlashAvatar in terms of LPIPS, SSIM, and PSNR metrics across all three subjects, achieving the best results in the shortest training time of only 8 minutes (on a single Nvidia A6000 GPU). We demonstrate qualitative expression reconstruction results in Figure~\ref{fig:imavatar_sota}, with the first column displaying source expressions that are subsequently applied to the learned head avatars.

\begin{figure}[h]
    \centering
    \includegraphics[width=\linewidth]{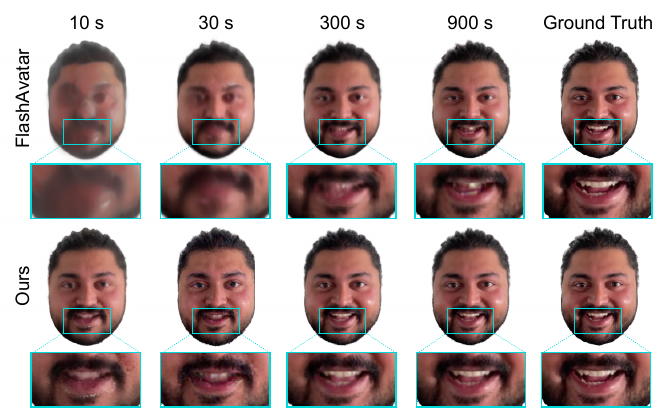}
    \caption{Convergence speed on video-based optimization.}
    \label{fig:converge-speed}
    \vspace{-0.25cm}
\end{figure}

We compare the convergence speed with FlashAvatar, which is known for its rapid training. Figure~\ref{fig:converge-speed} displays the comparison. Our approach, utilizing a better initialization and a non-neural network strategy, achieves faster convergence than FlashAvatar. Ultimately, our method enables the efficient optimization of a high-density 3D Gaussian head avatar within minutes.

\subsection{Ablation Studies}

\begin{figure}[h]
    \centering
    \includegraphics[width=\linewidth]{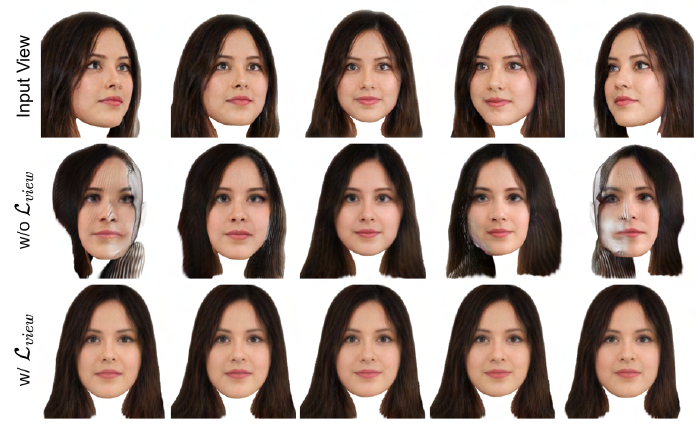}
      \caption{Ablation results on view consistency regularization.}
    \label{fig:ablation-imavatar-reg-view}
\end{figure}

We first use an example to demonstrate the importance of our view consistency regularization loss $\mathcal{L}_{\text{view}}$, see Figure~\ref{fig:ablation-imavatar-reg-view}. The first row displays input images of the same identity from different camera views fed into our single-image-based reconstruction model. The second (w/o $\mathcal{L}_{\text{view}}$) and third (w/ $\mathcal{L}_{\text{view}}$) rows show the reconstructed heads rendered from a frontal view. As evident, without $\mathcal{L}_{\text{view}}$, the reconstruction quality is biased towards the side visible in the input image. However, with $\mathcal{L}_{\text{view}}$, the reconstructions appear nearly identical across different views of the same identity, showcasing the robustness of our approach in maintaining consistent visual outputs.

\begin{figure}[h]
    \centering
    \includegraphics[width=\linewidth]{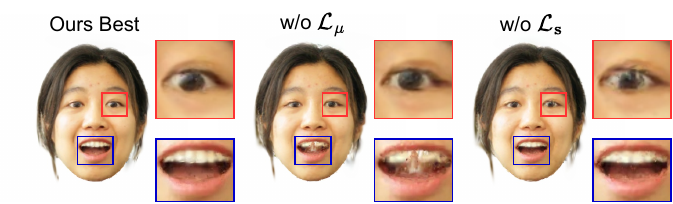}
      \caption{Ablation results on position and scale regularizations.}
    \label{fig:ablation-imavatar-reg-pos-scale}
\end{figure}

We demonstrate the critical role of our 3D Gaussian position and scale regularization losses in Figure~\ref{fig:ablation-imavatar-reg-pos-scale}. In this experiment, we trained two additional single-image-based reconstruction models, each omitting either the $\mathcal{L}_{\mu}$ (position loss) or $\mathcal{L}_{\mathbf{s}}$ (scale loss). These models were used to initialize 3D Gaussian heads, which were then further optimized using video data. The results clearly show that without adequate regularization of the position and scale of the 3D Gaussians, noticeable artifacts persist even after subsequent video-based optimization.

\begin{figure}[h]
    \centering
    \includegraphics[width=\linewidth]{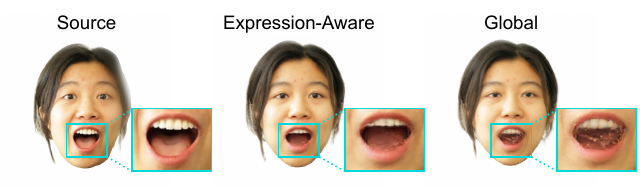}
    \caption{Ablation results on different levels of rectification.}
    \vspace{-0.25cm}
    \label{fig:ablation-imavatar}
\end{figure}

In Figure~\ref{fig:ablation-imavatar}, we showcase the importance of our expression-aware blendmaps. We apply the expression estimated from the source image to our learned head avatars. The second image displays the head enhanced with our expression-aware UV blendmaps and the third illustrates the face with only global UV rectification. Both approaches use the same amount of time for optimization in this ablation experiment. It becomes clear that global UV rectification introduces artifacts when dealing with exaggerated expressions. Quantitative results, which further substantiate the limitations of using only global rectification, are presented in Table~\ref{table:imavatar_compare}.

For more ablation study results please refer to our supplementary material.

\section{Conclusion}
\label{sec:conc}

In this work, we propose Déjà-vu, a novel framework designed to create controllable 3D Gaussian head avatars with fast training. Notably, Déjà-vu is the first method capable of reconstructing a 3D Gaussian head based solely on a single image as input and trained using 2D images. This reconstruction model provides robust initialization for our video-based optimization. We propose learnable UV blendmaps to adjust the head to have the desired expression, a solution that is both effective and quick to train. Our framework outperforms existing state-of-the-art methods in both rendering quality and training speed. In the future, we will enhance the adaptability of Déjà-vu to a wider range of facial expressions and explore more applications.

\newpage

{\small
\bibliographystyle{ieee_fullname}
\bibliography{references}
}

\end{document}